\documentclass[letterpaper, 10 pt, conference]{ieeeconf}  % Comment this line out if you need a4paper

\IEEEoverridecommandlockouts                              % This command is only needed if 
\title{\LARGE \bf
TransParking: A Dual-Decoder Transformer Framework with Soft Localization for End-to-End Automatic Parking
}

\author{Hangyu Du, Chee-Meng Chew$^{*}$%
\thanks{All authors are with the College of Design and Engineering, National University of Singapore, Singapore. Emails: {\tt\small e1373656@u.nus.edu, mpeccm@nus.edu.sg}. $\ast$ is the corresponding author.}%
}

\usepackage{cite}
\usepackage{amsmath, amssymb}
\usepackage{graphicx}
\usepackage[margin=1in]{geometry}
\usepackage{booktabs}   
\usepackage{graphicx}   
\usepackage{amsmath}    
\usepackage{amssymb}    
\begin{document}

\maketitle
\thispagestyle{empty}
\pagestyle{empty}

%%%%%%%%%%%%%%%%%%%%%%%%%%%%%%%%%%%%%%%%%%%%%%%%%%%%%%%%%%%%%%%%%%%%%%%%%%%%%%%%
\begin{abstract}
In recent years, fully differentiable end-to-end autonomous driving systems have become a research hotspot in the field of intelligent transportation. Among various research directions, automatic parking is particularly critical as it aims to enable precise vehicle parking in complex environments. In this paper, we present a purely vision-based transformer model for end-to-end automatic parking, trained using expert trajectories. Given camera-captured data as input, the proposed model directly outputs future trajectory coordinates. Experimental results on the ParkingE2E real-vehicle dataset show that our method reduces the Hausdorff distance from 0.386 m to 0.2156 m, the L2 distance from 0.1193 m to 0.06296 m, and the Fourier descriptor difference from 2.431 to 1.239, under identical settings. Our approach thus provides an effective solution for fully differentiable automatic parking.

\end{abstract}

%%%%%%%%%%%%%%%%%%%%%%%%%%%%%%%%%%%%%%%%%%%%%%%%%%%%%%%%%%%%%%%%%%%%%%%%%%%%%%%%
\section{INTRODUCTION}

The traditional automatic parking systems are usually composed of three modules: perception, planning and control. In these systems, the vehicle initially employs sensors such as cameras and LiDAR to detect and model its surrounding environment, identifying both the target parking space and any nearby obstacles. Subsequently, a trajectory is planned based on the acquired perception data, and the vehicle’s speed and direction are continuously adjusted in real time—often via a PID controller or other conventional control algorithms to execute the parking maneuver. Although many of these algorithms have been successfully deployed in real-world vehicles, forming a mature industrial ecosystem, they remain highly dependent on accurate environmental information and precise vehicle dynamics models. As a result, perception errors can be amplified as they propagate through planning and control, increasing the risk of suboptimal maneuvers in complex or occluded environments.

\begin{figure*}[t!]
  \centering
  \includegraphics[width=\textwidth]{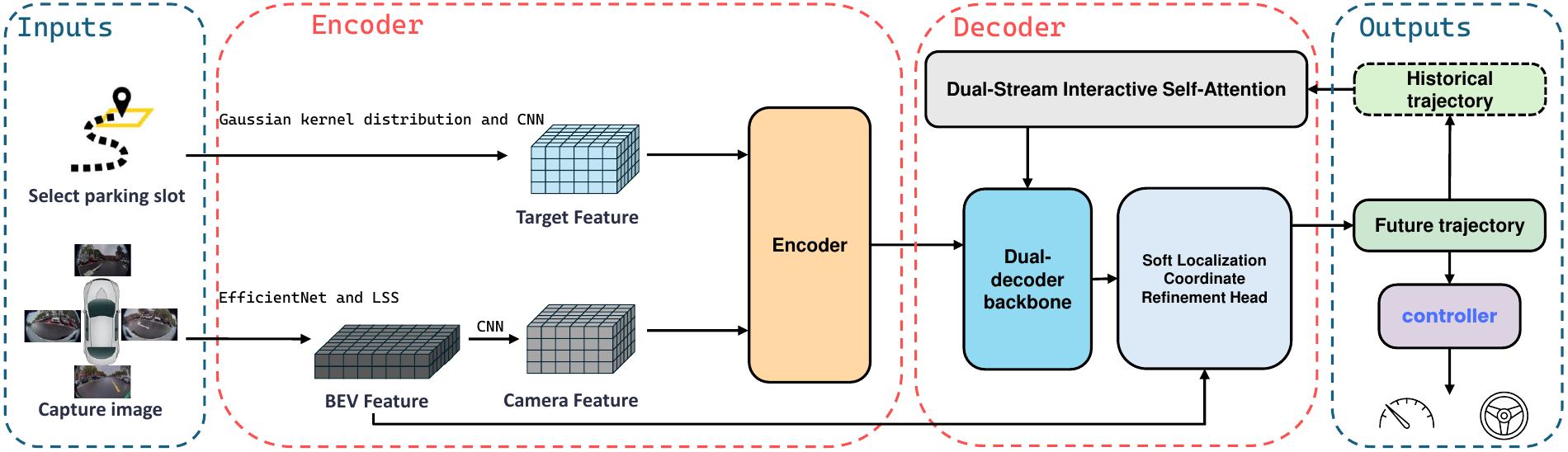} 
  \caption{Overall architecture. Given the user-selected slot and multi-camera images, we form a Gaussian soft query and build BEV features via EfficientNet+LSS. Dual decoders predict $(x_t,y_t)$ with Dual-Stream Interactive Self-Attention, and a Soft Localization head refines each step using local BEV evidence before outputting the final trajectory (used by the controller).}

  \label{P1}
\end{figure*}

Inspired by the Transformer model\cite{vaswani2017attention}, this paper proposes a pure vision-based end-to-end autonomous driving approach, effectively mitigating the aforementioned problems. Prior end-to-end parking pipelines often feed x- and y-coordinates alternately into a single decoder, which stretches the sequence and breaks the simultaneity of $(x_t,y_t)$ that holds in real vehicle motion, thereby accumulating errors. We therefore employ two decoders that produce $x_t$ and $y_t$ at the same time step and let them interact through Dual-Stream self-attention, yielding shorter sequences and stronger $x$–$y$ coupling.
 The specific contributions of this work are as follows: 

\begin{itemize}

\item We examined the model architectures proposed in previous end-to-end automatic parking studies and found that sequentially inputting the coordinates of x and y into a single decoder does not fully adhere to the real-world temporal order of coordinate inputs, which leads to greater trajectory errors.
\item As shown in Figure~\ref{P1}, we propose an end-to-end neural network model \textbf{based on a dual-decoder structure}. comparative experiments demonstrate that our approach achieves more accurate and robust future trajectory predictions compared to earlier end-to-end automatic parking solutions.

\end{itemize}

\section{RELATED WORK}

\subsection{BEV representations for autonomous driving}
Bird’s-Eye View (BEV) representations offer a clear, top-down perspective, showing vehicles, roads, and obstacles at their true planar positions. Consequently, BEV is widely adopted in autonomous driving applications. However, cameras cannot directly acquire distance information for each pixel. To address this, LSS \cite{philion2020lift} compensates for the lack of depth information by predicting the depth distribution of each pixel. BEVDepth~\cite{li2023bevdepth}, on the foundation of LSS, incorporates explicit depth supervision and a Depth Refinement Module, enhancing the reliability of depth estimation. BEVDet\cite{huang2021bevdet} was improved on the basis of LSS and achieved higher accuracy. BEVDet4D~\cite{huang2022bevdet4d}, built on BEVDet, introduces a time dimension to fuse spatiotemporal features. BEVFormer\cite{li2022bevformer} is a Transformer-based BEV feature generation algorithm. It can simultaneously integrate the spatial and temporal information of multi-view images and reduce the computational load, generating high-quality features. DualBEV\cite{li2024dualbev}  introduces a dual feature fusion strategy to effectively predict BEV probabilities. BEVFusion\cite{liu2023bevfusion} implements a multimodal fusion perception scheme that combines LiDAR and image data, effectively preserving both geometric and semantic information. FastBEV~\cite{li2024fast} proposed Fast-Ray, thereby significantly enhancing the inference speed. These BEV methods motivate our choice of LSS to obtain stable camera-to-BEV features. Unlike the binary slot query used in ParkingE2E~\cite{li2024parkinge2e}, we adopt a Gaussian soft query that provides continuous spatial weights and smoother gradients within the target region (see Fig.~\ref{P2}).

\subsection{End-to-End Autonomous Driving}
In recent years, neural networks have substantially advanced end-to-end autonomous driving~\cite{bojarski2016end}. For clarity, we group prior work into city-driving and parking scenarios and review representative approaches in each.

\noindent\textbf{Urban driving:} Early end-to-end approaches directly map front-view images and commands to controls~\cite{codevilla2018end}, while later methods inject trajectory supervision~\cite{wu2022trajectory} or fuse multimodal inputs with transformers~\cite{chitta2022transfuser,jaeger2023hidden,shamsoshoara2024swaptransformer,mu2024pix2planning,lyu2024sensor}. Despite strong results in city driving, many pipelines either predict controls directly or predict 2D waypoints/trajectories, and are not specifically tailored to the low-speed, high-precision geometric constraints of parking.

\noindent\textbf{Parking:} Parking-focused works include two-stage optimization~\cite{chai2018two}, reinforcement learning~\cite{zhang2019reinforcement}, and end-to-end pipelines that map images to controls or trajectories~\cite{shen2020parkpredict,shen2022parkpredict+,chai2022deep,yang2024e2e,li2024parkinge2e}. We follow the ``images~$\rightarrow$~trajectories'' route but differ from prior work by (i) synchronously decoding $(x_t,y_t)$ with dual decoders, (ii) enabling explicit cross-stream interaction, and (iii) applying a soft-localization refinement that leverages local BEV context before emitting the final coordinates.

\begin{figure*}[t!]
  \centering
  \includegraphics[width=\textwidth]{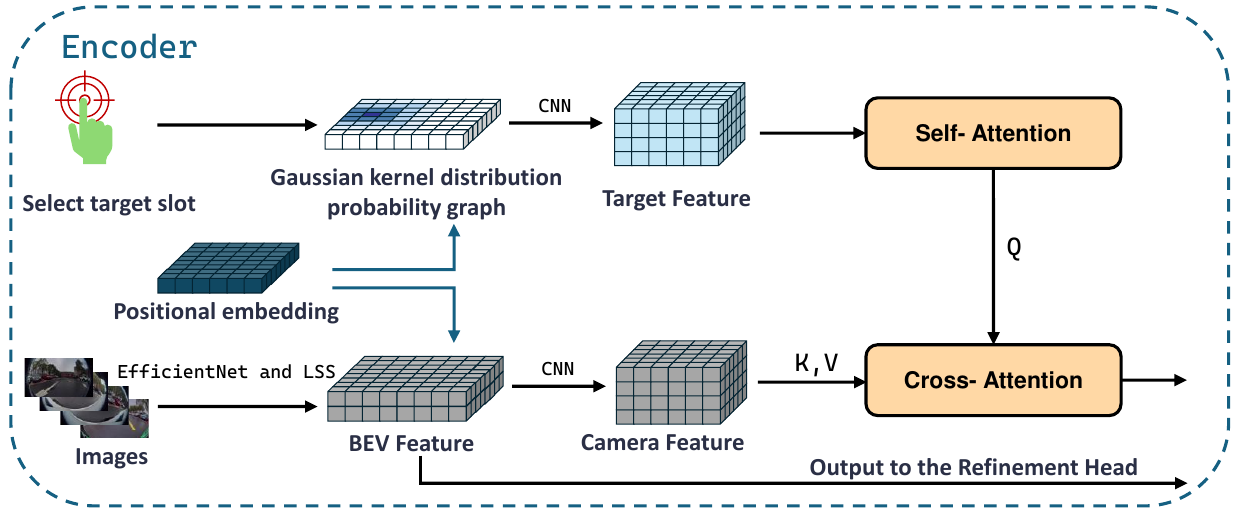} 
  \caption{The specific implementation of the Encoder structure}
  \label{P2}

\end{figure*}

\noindent\textbf{Most relevant baseline:}
ParkingE2E~\cite{li2024parkinge2e} generates a binary map around the selected slot and convolves it to form a query vector that attends to BEV features to predict trajectory points. While this explicit spatial prior highlights the target area, it offers no gradation within the region and creates sharp boundaries that may reduce attention continuity. We replace it with a Gaussian soft query and further add a soft-localization refinement head (see Fig.~\ref{P2} and Fig.~\ref{P4}).

\section{METHODOLOGY}

\subsection{Problem Description and Optimization Goals} 
We conduct training via an end-to-end neural network and define the dataset as:
\begin{equation}
\mathcal{D} = \left\{ \left( I_{i,j}^k,\, X_{i,j},\, Y_{i,j},\,S_i,\, G_{i,j} \right) \right\},
\end{equation}
where trajectory index $i \in [1, M]$, trajectory points index $j \in [1, N_i]$, camera index $k \in [1, R]$, RGB image $I$, ground truth trajectory point $X$ and $Y$, 2D target slot coordinates $S$, The BEV weight probability map $G$ based on Gaussian kernel distribution generated by the ground truth trajectory point $X$ and $Y$. Reorganize the dataset as:
\begin{equation}
\mathcal{X}^{\prime}_{i, j}=\left\{X_{i, \min \left(j+b, N_{i}\right)}\right\}_{b=1, 2, \ldots, Q,}
\end{equation}
\begin{equation}
\mathcal{Y}^{\prime}_{i, j}=\left\{Y_{i, \min \left(j+b, N_{i}\right)}\right\}_{b=1, 2, \ldots, Q,}
\end{equation}
\begin{equation}
\mathcal{M}_{i, j}=\left\{G_{i, \min\left(j+b, N_{i}\right)}\right\}_{b=1,2,\ldots,Q,}
\end{equation}
and
\begin{equation}
\mathcal{D}^{\prime}=\left\{\left(I_{i, j}^{k},\, \mathcal{X}^{\prime}_{i, j},\, \mathcal{Y}^{\prime}_{i, j},\,S_i, \, \mathcal{M}_{i, j}\right)\right\},
\end{equation}
where Q denotes the length of the predicted trajectory points.

The optimization goals for the end-to-end network are as follows:
\begin{equation}
\begin{gathered}
\theta^{*} = \underset{\theta}{\operatorname*{arg\,min}}\, \mathbb{E}_{\left(I,\mathcal{X}',\mathcal{Y}',S,\mathcal{M}\right) \sim \mathcal{D}'} \bigl[ \mathcal{L}_{x}\bigl(\mathcal{X}',\mathcal{N}^{x}_{\theta}(I,S)\bigr) \\
\quad + \mathcal{L}_{y}\bigl(\mathcal{Y}',\mathcal{N}^{y}_{\theta}(I,S)\bigr)
+ \mathcal{L}_{m}\bigl(\mathcal{M},\mathcal{N}^{p}_{\theta}(I,S)\bigr) \bigr],
\end{gathered}
\end{equation}
where \(\mathcal{L}\) denotes the loss function, \(\mathcal{N}^{x}_{\theta}(I,S)\), \(\mathcal{N}^{y}_{\theta}(I,S)\) and \(\mathcal{N}^{p}_{\theta}(I,S)\) respectively denote the predicted X-coordinates, Y-coordinates and two-dimensional weight probability map of the output of the model.

\subsection{End-to-End Neural Network Model Based on Camera}

1) Encoder: 

\textbf{Query:} In ParkingE2E\cite{li2024parkinge2e}, the authors first generate a binary map with the same spatial dimensions as the BEV feature map and a depth of 1. In this map, the target slot and its surrounding fixed region are marked with a value of 1, while the area outside this range is marked with 0. After applying convolutional neural network, this map is used as a query vector to compute attention with the BEV feature map, serving as the final output of the encoder. This approach offers us significant inspiration and insight. The method of ParkingE2E clearly indicates to the model which areas need attention, However, a drawback is that it does not distinguish subregions within the target area, and the abrupt boundary transition might adversely affect continuity in attention. Therefore, in our work, we aim to give greater weight to the center while retaining some information from the surrounding areas, and this weight gradually weakens as the distance increases. We can select the desired parking slot by using the interactive tool ‘2D Nav Goal' of Rviz and send the coordinates to the model. Subsequently, a zero tensor with the same width and height as the BEV feature map and a depth of 1 is created. In this channel, a probability map following a Gaussian kernel distribution is generated within the specified range centered on the coordinates of target slot. Subsequently, the tensor undergoes feature extraction via same convolutional neural network, transforming it into a high-dimensional vector. The formula is as follows:  
\begin{equation}
G(i, j)=\exp \left(-\frac{(i-c x)^{2}+(j-c y)^{2}}{2 \sigma^{2}}\right),
\end{equation}
where $cx, cy$ represent the pixel coordinates of the target slot, $i, j$ represents the pixel coordinates on the BEV image where the Gaussian values need to be calculated. $\sigma$ represents the standard deviation of the Gaussian distribution.

As shown in Table~\ref{T3}, our ablation experiments demonstrate that the proposed method not only ensures strict selection of the target area but also enhances the granularity within the target area. This clear signal is beneficial for subsequent network training and backpropagation, thereby promoting model convergence and feature strengthening.

\textbf{Key and Value:} We employ EfficientNet~\cite{tan2019efficientnet} to extract image features \(F_{img} \in \mathbb{R}^{C \times H_{img} \times W_{img}}\) from the RGB input, and then use the LSS\cite{philion2020lift} method to generate the BEV feature map. First, we predict the depth distribution for each pixel and map it to the 3D space. Then, we fuse the depth information with the image features. Finally, we project the fused features onto the BEV voxel grid using the camera extrinsics and intrinsics to generate the camera features. For the obtained feature map, the range in the \(x\)-direction is denoted as \([-R_x, R_x]\) m, where m indicates meters, and the range in the \(y\)-direction is denoted as \([-R_y, R_y]\) m.

\textbf{Final Output:} We first conduct self-attention computation on the high-dimensional vector Query. Then, we use the BEV feature map as Key and Value. Finally, cross-attention calculation is performed among Query, Key, and Value to obtain the fused feature vector. The overall calculation process is illustrated in Figure~\ref{P2}.

2) Decoder: 
In this paper, we treat trajectory prediction as a label classification task and optimize it through cross-entropy loss.

The sequence of the trajectory: In previous works, the common practice was to feed the X-coordinate and Y-coordinate sequentially into a decoder following the temporal order. Although this method is straightforward, yet a concealed issue exists. In the real world, the movement of a vehicle is associated with the simultaneous alteration of its X-coordinate and y-coordinate. For instance, Regarding Y-coordinate $Y_{t}$ at the time step t, its preceding time step ought to be $Y_{t-1}$. If we desire to pay additional attention to the information of X-coordinate, then $X_{t-1}$ can also be regarded as the previous time step. In any case, it should not be $X_{t}$. Therefore, treating the X-coordinate and Y-coordinate at the same moment as two separate steps is not only a logic that makes it difficult for the model to understand, but also the longer sequence generated will lead to more error accumulation.

We separate the X and Y coordinates of the model in the ParkingE2E~\cite{li2024parkinge2e}, and then train and predict using two decoders with the same number of layers. At each time step, the model simultaneously outputs the X and Y coordinates, and they can only see the historical information of the same type of coordinates. Based on the findings of the comparative experiments depicted in Table~\ref{T1}, the experimental outcome of jointly determining the most appropriate X-coordinate and Y-coordinate at one time based on the perceived surrounding environmental information surpasses the approach of splitting the X-coordinate and Y-coordinate into two time steps.

In this paper, Trajectory serialization represents trajectory points as discrete tokens. By serializing the trajectory points, the position regression can be turned into token prediction. Subsequently, we can leverage a transformer decoder to predict the trajectory point \((p^x_{i,j}, p^y_{i,j})\) in the ego vehicle's coordinate system. We utilize the following serialization method:
\begin{equation}
\mathrm{Ser}(p^x_{i,j}) = \left\lfloor \frac{p^x_{i,j} + R_x}{2R_x} \right\rfloor \times N_t,
\end{equation}

\begin{equation}
\mathrm{Ser}(p^y_{i,j}) = \left\lfloor \frac{p^y_{i,j} + R_y}{2R_y} \right\rfloor \times N_t,
\end{equation}

where \(N_t\) represents the maximum value that can be encoded by a token in the sequence and the symbol for serializing trajectory points is denoted as \(\mathrm{Ser}(\cdot)\). \(R_x\) and \(R_y\) represent the maximum values of the predicted range in the \(x\) and \(y\) directions, respectively.

Ultimately, the trajectory sequences of the X and Y coordinates fed into the decoder are respectively: 

\begin{equation}
\begin{array}{c}
{\left[\operatorname{BOS}, \operatorname{Ser}\left(P_{i, 1}^{x}\right), \operatorname{Ser}\left(P_{i, 2}^{x}\right),\ldots,\right.} \\
\left.\operatorname{Ser}\left(P_{i, N_{i}}^{x}\right), \operatorname{EOS}\right],
\end{array}
\end{equation}

\begin{equation}
\begin{array}{c}
{\left[\operatorname{BOS}, \operatorname{Ser}\left(P_{i, 1}^{y}\right), \operatorname{Ser}\left(P_{i, 2}^{y}\right),\ldots,\right.} \\
\left.\operatorname{Ser}\left(P_{i, N_{i}}^{y}\right), \operatorname{EOS}\right],
\end{array}
\end{equation}
where BOS represents the start flag and EOS represents the end flag.

\begin{table}[htbp]
    \centering
    \caption{Evaluation of Comparative Performance}
    \label{T1}
    \resizebox{0.49\textwidth}{!}{%
        \begin{tabular}{lccc}
            \toprule
            \textbf{Method} & \textbf{Haus. Dis. (m) \(\downarrow\)} & \textbf{L2 Dis. (m) \(\downarrow\)} & \textbf{Four. Diff.} \\
            \midrule
            \textbf{ParkingE2E-Split} & \textbf{0.2967} & \textbf{0.0808} & \textbf{1.904} \\
            ParkingE2E\cite{li2024parkinge2e}       &  0.386 & 0.1193 & 2.431 \\
            \bottomrule
        \end{tabular}
    }
\end{table}
\textbf{Decoder backbone:} The BEV features serve as the key and value, while the serialization sequence is used as the query. The hidden states of the X and Y coordinates of the future trajectory are generated in an autoregressive method by the dual-decoder architecture. During training, we add position embedding to the sequence points and use a mask matrix to achieve parallelization.

\begin{figure}[t!]
  \centering
  \includegraphics[width=\columnwidth]{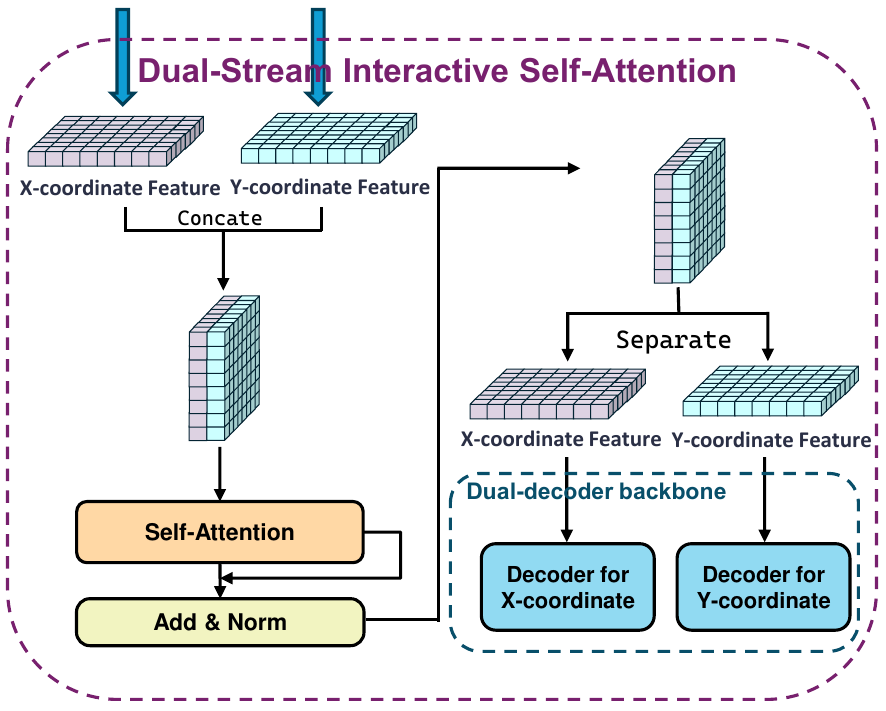} 
  \caption{The implementation logic of the Dual-Stream Interactive Self-Attention structure. The feature vectors of the X and Y coordinates are concatenated at each time step, that is, there are two vectors at each time step. After the attention calculation is completed, it is reshaped to the original size. The Batch dimension is ignored in the picture.}
  \label{P3}
\end{figure}
\begin{figure*}[t!]
  \centering
  \includegraphics[width=0.7\textwidth]{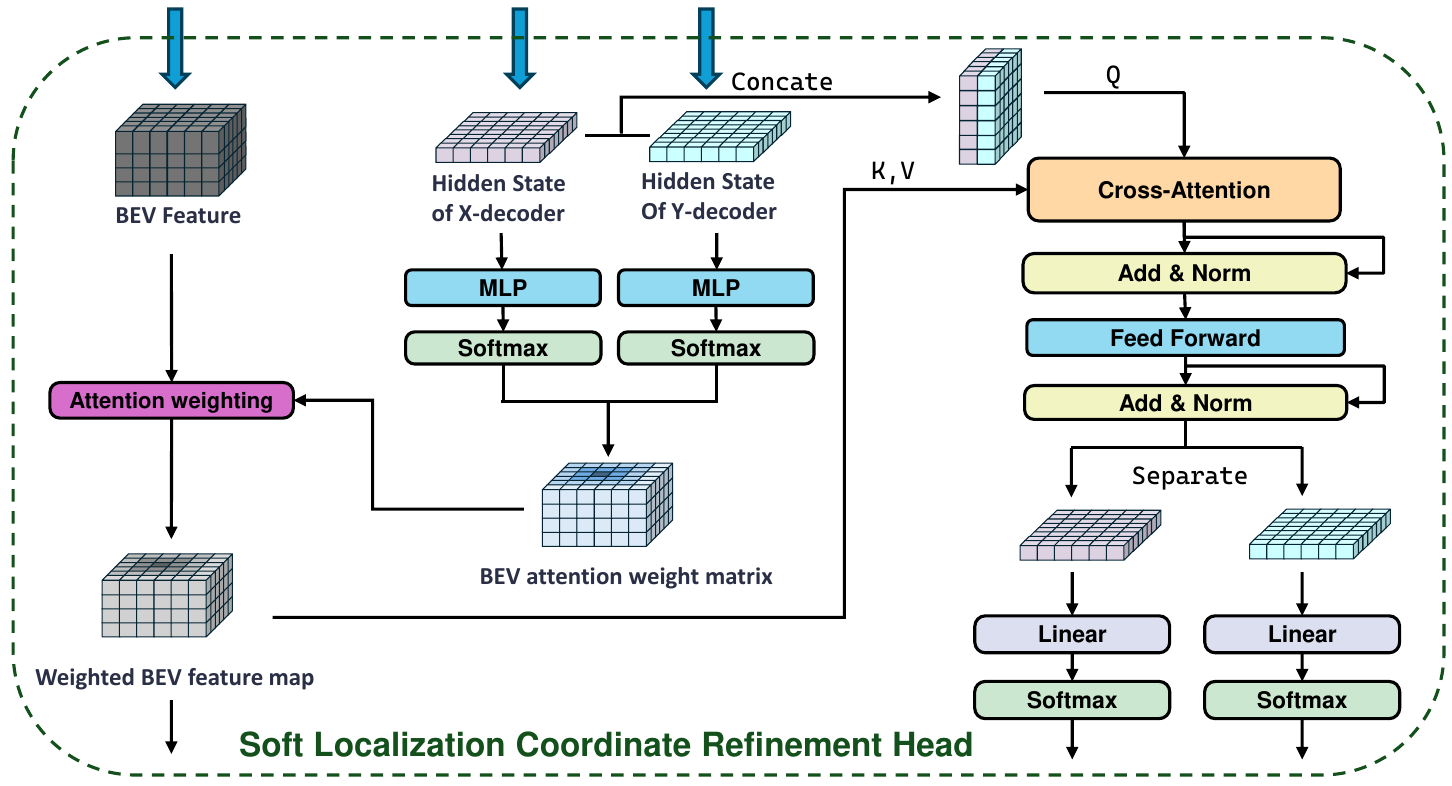}
  \caption{The structure of the Soft Localization Coordinate Refinement Head. The BEV feature comes from the result of LSS. The hidden state vectors of the decoders of X and Y jointly predict the attention weight map of the possible area of the next trajectory point, which is multiplied by the BEV feature to generate the BEV feature weight map. The hidden state vectors interact with it through the cross-attention mechanism to achieve fine-tuning.}
  \label{P4}
\end{figure*}
\textbf{Dual-Stream Interactive Self-Attention:} For trajectory prediction models, the X and Y coordinates usually have a close correlation. However, if only a dual-decoder structure is used for token prediction of trajectory points, it is still difficult to capture the coupling relationship between the X-coordinate and Y-coordinate. In such circumstances, during the self-attention calculation, $X_{t}$ coordinate  only attends to the X-coordinate from $X_{0}$ to $X_{t-1}$, ignoring Y-coordinate features in the same period. Likewise, $Y_{t}$ cannot utilize from $X_{0}$ to $X_{t-1}$. 

To address the deficiency that the X and Y coordinates have no knowledge of each other's historical information, we propose to initially concatenate the feature vectors of the X and Y coordinates at each time step, then conduct a self-attention mechanism calculation and obtain the fused feature vector. Subsequently, the fused feature vector is separated into x and y, which are respectively input into their respective decoders. The specific structure is presented as shown in the Figure~\ref{P3}. This attention module ensures that the high-dimensional feature vectors of the X and Y coordinate at the same time step are capable of integrating each other's feature information. It guarantees that, in the subsequent mask self-attention layers in their respective decoders, The $X_{t}$ and $Y_{t}$ can indirectly take into account the X-coordinate and Y-coordinate information from $0$ to $t-1$.

\textbf{Soft Localization Coordinate Refinement Head:} A considerable number of previous related endeavors have lacked attention to the environmental information surrounding the future trajectory points predicted by vehicles. Hence, we expect that before the model computes the final position of the next trajectory point, it should initially inspect the environmental characteristics of this area and make appropriate fine-tuning based on the environmental feature vector, and subsequently output the ultimate coordinates. Benefiting from the dual-decoder structure, We can predict the X and Y coordinates at the same time step simultaneously, which furnishes the requisite conditions to achieve this objective.

Specifically, for the hidden state vectors of the X and Y coordinates output by the dual-decoder, after passing through the MLP layer, a learnable BEV attention weight matrix of the same size as the BEV feature map is predicted. This matrix has a predicted center point, representing the approximate position of the next trajectory point. The closer to this central point, the higher the weight. As the distance increases, the weight decreases, typically following a Gaussian kernel distribution. The specific formula is:
\begin{equation}
G(i, j)=\frac{\exp \left(-\frac{(i-\mathrm{px})^{2}+(j-\mathrm{py})^{2}}{2 \sigma^{2}}\right)}{\sum_{u=0}^{h-1} \sum_{v=0}^{w-1} \exp \left(-\frac{(u-\mathrm{px})^{2}+(v-\mathrm{py})^{2}}{2 \sigma^{2}}\right)},
\end{equation}
where $px, py$ represent the coordinates of the Central pixel point, $i,j$ represents the pixel coordinates on the BEV image where the Gaussian values need to be calculated. $\sigma$ represents the standard deviation of the Gaussian distribution, $h,w$ represent the height and width of the BEV image.

Then, the element-wise multiplication of the BEV attention weight matrix and the BEV feature map yields the Weighted BEV feature map, which serves as the key and value for the cross-attention calculation with the hidden state vectors of the X and Y coordinates at the current time step. Subsequently, after undergoing Feedforward layer processing, residual connection, and normalization, the final coordinates are output. In this manner, at each time step, the model is prompted to contemplate how to output more precise coordinates based on the environmental features of the region, thereby preventing the model from the issue of farsightedness. Note that the task of the Weighted BEV feature map is to provide environmental features near the trajectory points for the X and Y coordinates in the cross-attention calculation. It will also be output eventually, with the aim of participating in the calculation of the loss function during the training phase. The specific steps are shown in the Figure~\ref{P4}.

\subsection{Lateral control and longitudinal control}

In our work, we adopt the similar approach as ParkingE2E\cite{li2024parkinge2e} for Lateral and Longitudinal Control of non-omnidirectional vehicle. During the control process, we mark the initial moment of parking as $t_{0}$. The predicted path $\mathcal{X}_{t_{0}}, \mathcal{Y}_{t_{0}} = \mathcal{N}_{\theta}(I_{t_0}, S)$ obtained at the initial moment is used. The relative pose from the initial moment $t_{0}$ to the current moment t can be obtained through the positioning system, denoted as  $ego_{t_0 \rightarrow t}$. The target steering angle  $\mathcal{A}^{tar}$ is obtained using the RWF (Rear-wheel Feedback) method, and its expression is as follows:

\begin{equation}
\mathcal{A}_{t}^{t a r}=\operatorname{RWF}\left(\mathcal{X}_{t_{0}}, \mathcal{Y}_{t_{0}},  e g o_{t_{0} \rightarrow t}\right).
\end{equation}
We achieve lateral and longitudinal control by employing a cascaded PID controller in accordance with the speed feedback $\mathcal{V}^{feed}$ and steering feedback $\mathcal{A}^{feed}$ from the chassis, as well as the target speed $\mathcal{V}^{tar}$ from the setting and the target steering $\mathcal{A}^{tar}$ from the calculation.

In this approach, the controller determines how the vehicle moves based on the future trajectory, vehicle posture and feedback signals, and only needs to focus on how to plan the route from this new starting point to the target position, without the need to constantly monitor the global positioning of the vehicle.
\begin{figure}[t!]
  \centering
  \includegraphics[width=\columnwidth]{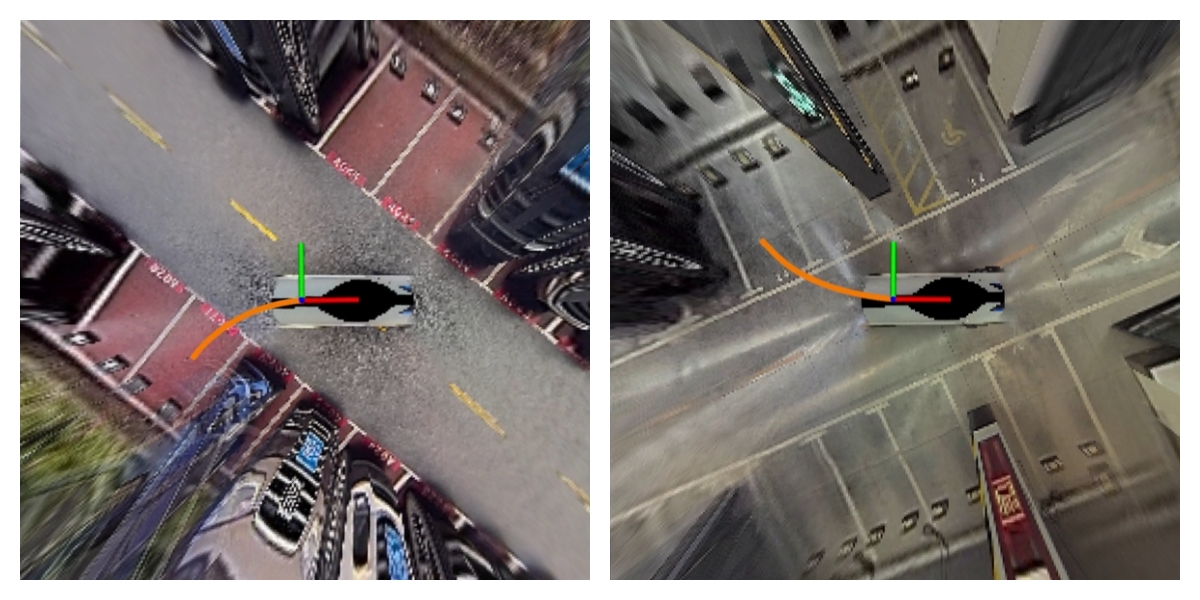} 
  \caption{The inference results in indoor and outdoor scenarios. The orange line represents the inference trajectory of the method proposed in this paper, and the purple line represents that of ParkingE2E~\cite{li2024parkinge2e}.}
  \label{E3}
\end{figure}

\section{EXPERIMENTS}

\subsection{The construction of the dataset}

The dataset of this work originates from the publicly available real-vehicle dataset offered by ParkingE2E\cite{li2024parkinge2e}, which does not contain sudden obstacles and negative samples. This dataset acquires RGB images around the vehicle through on-board cameras and employs sensor data fusion algorithms to achieve precise vehicle positioning. Meanwhile, it provides the two-dimensional coordinates of the parking target slot in the ego coordinate system for each time step.  

\subsection{Evaluation Metrics}
We evaluate the inference capacity of the model by comparing the disparity between the predicted trajectory and the expert trajectory. Specifically, the following evaluation metrics are adopted:

\textbf{L2 Distance:} gauges the straight-line distance between the predicted trajectory and the ground-truth trajectory in the Euclidean space. 

\textbf{Hausdorff Distance:} For each point in the predicted trajectory and the ground-truth trajectory, calculate the distance to the nearest point in the other trajectory, and take the maximum value of the two sets of minimum distances. 

\textbf{Fourier Descriptor Difference:}  transforms each of the two trajectories into the frequency domain, serving to measure the disparity between the predicted trajectory and the ground-truth trajectory. 

\subsection{Comparative experiment}
To guarantee the fairness of the comparative experiments, the basic training parameters employed by all the models remain consistent. The size of the BEV Features is $200 \times 200$, The actual spatial range is $x \in [-10\mathrm{m}, 10\mathrm{m}]$, $y \in [-10\mathrm{m}, 10\mathrm{m}]$ with a resolution of $0.1$ meters.
In decoders, the maximum value of the trajectory serialization $N_t$ is $1200$. The trajectory decoder generates a sequence of predictions with a length of $30$. Our training equipment: One NVIDIA GeForce RTX 4090 GPU. The experimental outcomes of the ParkingE2E model\cite{li2024parkinge2e} proposed in previous works, the transfuser model (based on the GRU decoder)\cite{chitta2022transfuser}, and our proposed model are presented in Table~\ref{T2}. Under the same dataset and settings, our approach attains 0.2156/0.06296/1.239 on Hausdorff/L2/Fourier, compared with 0.386/0.1193/2.431 for ParkingE2E (Table~\ref{T2}).

\begin{table}[htbp]
    \centering
    \caption{Evaluation of Comparative Performance}
    \label{T2}
    \resizebox{0.49\textwidth}{!}{%
        \begin{tabular}{lccc}
            \toprule
            \textbf{Method} & \textbf{Haus. Dis. (m) \(\downarrow\)} & \textbf{L2 Dis. (m) \(\downarrow\)} & \textbf{Four. Diff.} \\
            \midrule
            \textbf{Our approach} & \textbf{0.2156} & \textbf{0.06296} & \textbf{1.239} \\
            ParkingE2E\cite{li2024parkinge2e}       &  0.386 & 0.1193 & 2.431 \\
            TransFuser\cite{chitta2022transfuser}       &  0.5293 & 0.3517 & 8.674 \\
            \bottomrule
        \end{tabular}
    }
\end{table}
\begin{figure*}[t!]
  \centering
  \includegraphics[width=\textwidth]{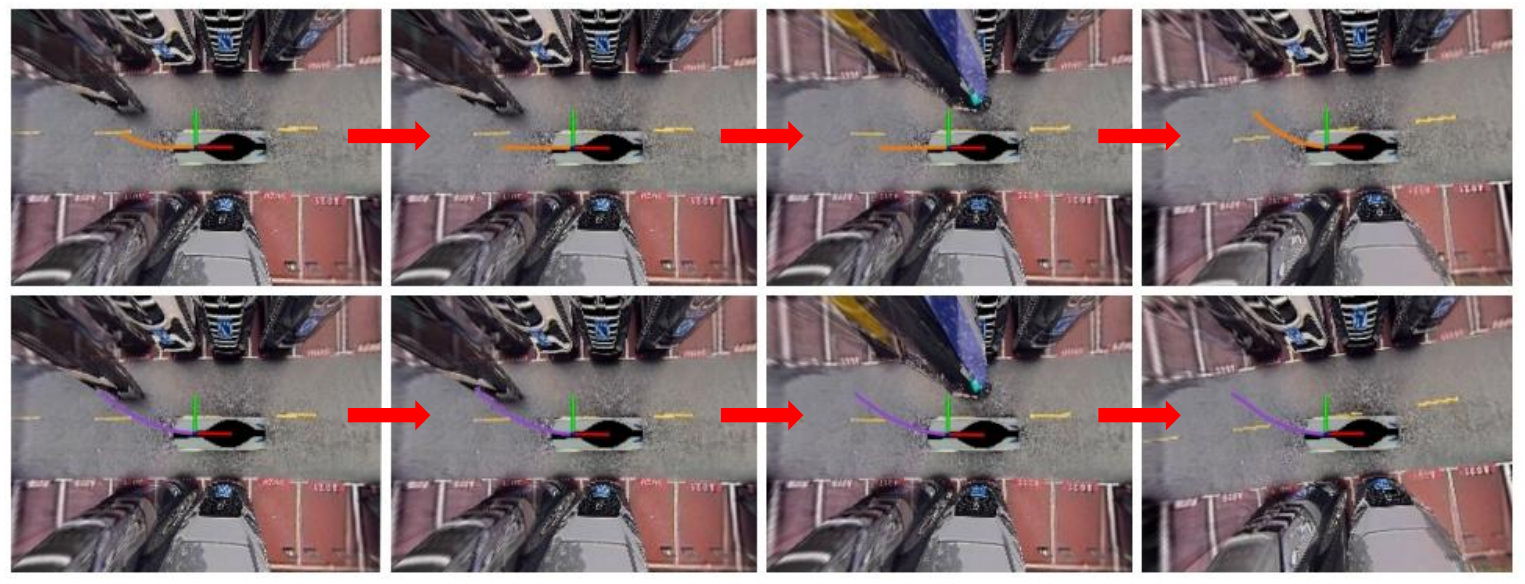} 
  \caption{The performance of the two algorithms in the same scenario where an obstacle suddenly appears. The orange line represents the inference trajectory of the method proposed in this paper, and the purple line represents that of ParkingE2E\cite{li2024parkinge2e}. Each row of images shows the trajectory inference results of four time steps from left to right.}
  \label{E1}
\end{figure*}
\subsection{Visualization and Analysis of inference}
\begin{figure}[t!]
  \centering
  \includegraphics[width=0.82\columnwidth]{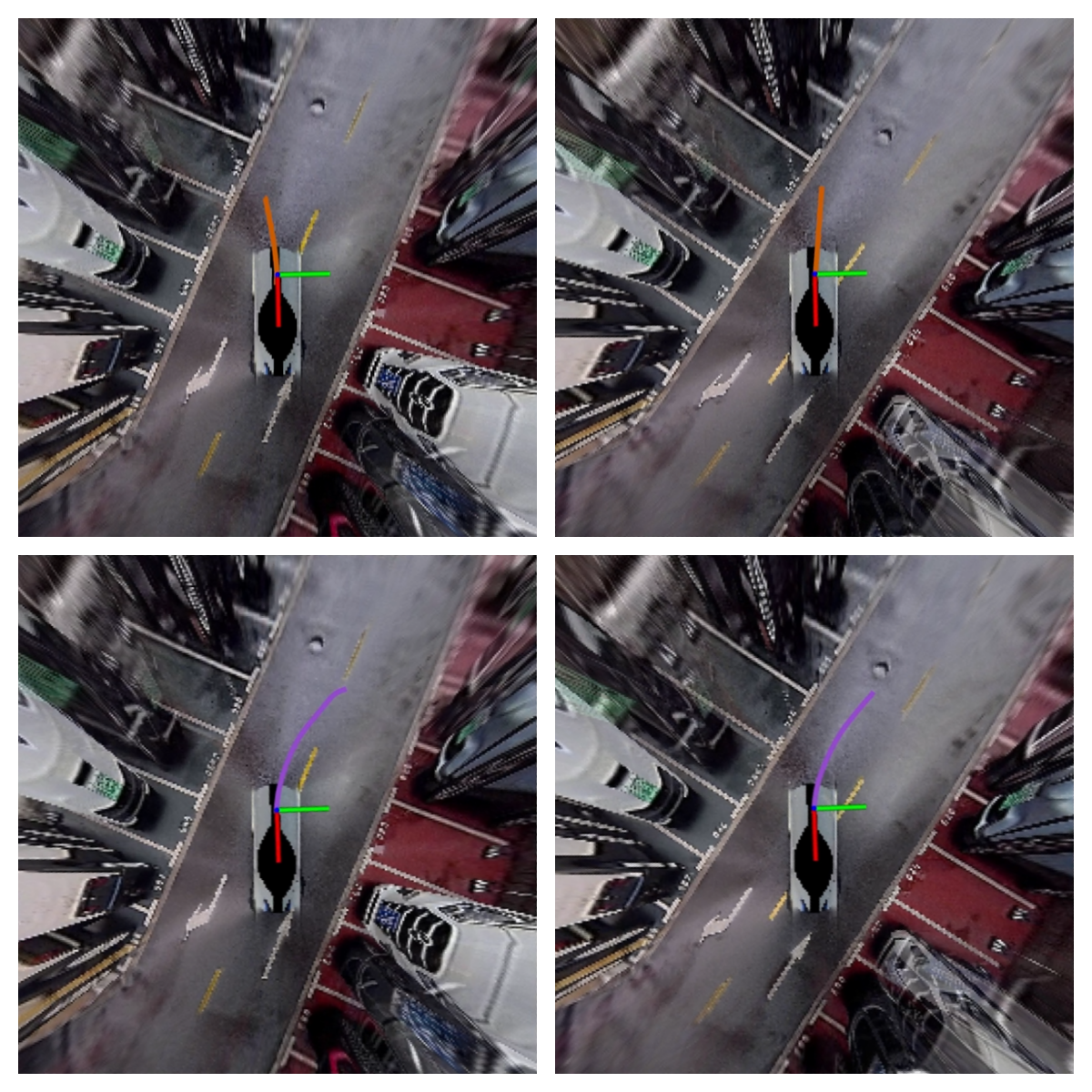} 
  \caption{The performance of the two algorithms in the same indoor scene. The orange line represents the inference trajectory of the method proposed in this paper, and the purple line represents that of ParkingE2E\cite{li2024parkinge2e}.}
  \label{E2}
\end{figure}

 We visualize the trajectory of model inference in Rviz, as shown in Figure~\ref{E3}. We also demonstrate that in the absence of special scenarios such as the sudden appearance of fast-moving objects in the training set, if other vehicles or pedestrians suddenly approach rapidly during the parking process, the model will immediately change its trajectory to prevent a collision. However, the model of ParkingE2E~\cite{li2024parkinge2e} mainly focuses its attention on the target area, thus being unable to modify the vehicle's driving trajectory in a timely manner, as shown in Figure~\ref{E1}. Meanwhile, we noticed that during the parking process, if the vehicle is still at a distance from the parking space and there are no obstacles, our model will first plan a relatively straight trajectory to avoid turning too early and crossing the parking line. Only when the vehicle approaches the parking space will it start to plan a curve to achieve the turn. Moreover, with the same number of Predicted token, the trajectory inferred by ParkingE2E tends to have a larger range and sometimes deviates from the route, inferring a tortuous path. The specific effect is shown in Figure~\ref{E2}. 

We analyzed the reason and found that the Soft Localization Coordinate Refinement Head in our model further constrains the local range through the attention heat map, making the decoded sequence more precise and convergent. Without this module, the model has less guidance or inhibition, higher degrees of freedom, and is more likely to lead to the sequence generation not converging to a shorter or more accurate trajectory. In autoregressive decoding, if the output range of the previous moment is too large or not precise enough, the subsequent predictions will accumulate errors, thereby causing this effect.

\subsection{Ablation experiments}
We conducted ablation experiments to verify the impact of each module. As shown in Table~\ref{T3}, we compared our complete model, the Decoder-Binary: dual-decoder structure using binary map target queries, the Decoder-Gaussian: dual-decoder structure using Gaussian target queries, the Decoder-Gaussian-Dual: dual-decoder structure using Gaussian target queries and adding Dual-Stream Interactive Self-Attention and the Decoder-Gaussian-Refinement: dual-decoder structure using Gaussian target queries and adding Soft Localization Coordinate Refinement Head. The experiment demonstrate that each module plays a certain role and our complete model achieves the best performance.  

\begin{table}[htbp]
    \centering
    \caption{Ablation experiments within the model}
    \label{T3}
    \resizebox{0.49\textwidth}{!}{%
        \begin{tabular}{lccc}
            \toprule
            \textbf{Method} & \textbf{Haus. Dis. (m) \(\downarrow\)} & \textbf{L2 Dis. (m) \(\downarrow\)} & \textbf{Four. Diff.} \\
            \midrule
            \textbf{complete model} & \textbf{0.2156} & \textbf{0.06296} & \textbf{1.239} \\
             Decoder-Binary   &  0.2967 & 0.08084 & 1.904 \\
             Decoder-Gaussian   &  0.2887 & 0.077 & 1.717 \\
             Decoder-Gaussian-Dual   &  0.2832 & 0.07556 & 1.665 \\
             Decoder-Gaussian-Refinement   &  0.2835 & 0.07092 & 1.631\\
            \bottomrule
        \end{tabular}
    }
\end{table}

\section{CONCLUSIONS}
In this paper, we identified the temporal issue in existing transformer-based trajectory prediction methods and addressed it by developing a dual-decoder architecture. Our experiments demonstrate that this approach improves accuracy and robustness. However, due to the limited scale and scenarios of the dataset, the current training results still have a gap compared with traditional methods. Therefore, we have only conducted trajectory inference on images collected from real vehicles so far and have not yet carried out road tests. In future work, we plan to expand the dataset, strengthen the model's capability of handling more complex tasks, and carry out real vehicle tests. The current query of encoder still determines a fixed range by parameters. In the future, we can make an attempt to employ reinforcement learning to allow the model to focus on a target range more flexibly.

 Overall, we are convinced that the learning-based trajectory prediction  is capable of continuous evolution, genuinely attaining and exceeding the level of traditional methods, and bringing a wider range of development space to the domain of autonomous driving.  

\addtolength{\textheight}{-12cm}   % This command serves to balance the column lengths

%%%%%%%%%%%%%%%%%%%%%%%%%%%%%%%%%%%%%%%%%%%%%%%%%%%%%%%%%%%%%%%%%%%%%%%%%%%%%%%%

%%%%%%%%%%%%%%%%%%%%%%%%%%%%%%%%%%%%%%%%%%%%%%%%%%%%%%%%%%%%%%%%%%%%%%%%%%%%%%%%

%%%%%%%%%%%%%%%%%%%%%%%%%%%%%%%%%%%%%%%%%%%%%%%%%%%%%%%%%%%%%%%%%%%%%%%%%%%%%%%%

\bibliographystyle{ieeetr}
\bibliography{sample}

@article{zhang2019reinforcement,
  title={Reinforcement learning-based end-to-end parking for automatic parking system},
  author={Zhang, Peizhi and Xiong, Lu and Yu, Zhuoping and Fang, Peiyuan and Yan, Senwei and Yao, Jie and Zhou, Yi},
  journal={Sensors},
  volume={19},
  number={18},
  pages={3996},
  year={2019},
  publisher={MDPI}
}

@article{vaswani2017attention,
  title={Attention is all you need},
  author={Vaswani, Ashish and Shazeer, Noam and Parmar, Niki and Uszkoreit, Jakob and Jones, Llion and Gomez, Aidan N and Kaiser, {\L}ukasz and Polosukhin, Illia},
  journal={Advances in neural information processing systems},
  volume={30},
  year={2017}
}

@article{bojarski2016end,
  title={End to end learning for self-driving cars},
  author={Bojarski, Mariusz and Del Testa, Davide and Dworakowski, Daniel and Firner, Bernhard and Flepp, Beat and Goyal, Prasoon and Jackel, Lawrence D and Monfort, Mathew and Muller, Urs and Zhang, Jiakai and others},
  journal={arXiv preprint arXiv:1604.07316},
  year={2016}
}

@inproceedings{philion2020lift,
  title={Lift, splat, shoot: Encoding images from arbitrary camera rigs by implicitly unprojecting to 3d},
  author={Philion, Jonah and Fidler, Sanja},
  booktitle={Computer Vision--ECCV 2020: 16th European Conference, Glasgow, UK, August 23--28, 2020, Proceedings, Part XIV 16},
  pages={194--210},
  year={2020},
  organization={Springer}
}

@article{li2022bevformer,
  title={BEVFormer: Learning Bird's-Eye-View Representation from Multi-Camera Images via Spatiotemporal Transformers},
  author={Li, Zhiqi and Wang, Wenhai and Li, Hongyang and Xie, Enze and Sima, Chonghao and Lu, Tong and Yu, Qiao and Dai, Jifeng},
  journal={arXiv e-prints},
  pages={arXiv--2203},
  year={2022}
}

@inproceedings{li2024dualbev,
  title={DualBEV: Unifying Dual View Transformation with Probabilistic Correspondences},
  author={Li, Peidong and Shen, Wancheng and Huang, Qihao and Cui, Dixiao},
  booktitle={European Conference on Computer Vision},
  pages={286--302},
  year={2024},
  organization={Springer}
}

@article{huang2021bevdet,
  title={Bevdet: High-performance multi-camera 3d object detection in bird-eye-view},
  author={Huang, Junjie and Huang, Guan and Zhu, Zheng and Ye, Yun and Du, Dalong},
  journal={arXiv preprint arXiv:2112.11790},
  year={2021}
}

@inproceedings{liu2023bevfusion,
  title={Bevfusion: Multi-task multi-sensor fusion with unified bird's-eye view representation},
  author={Liu, Zhijian and Tang, Haotian and Amini, Alexander and Yang, Xinyu and Mao, Huizi and Rus, Daniela L and Han, Song},
  booktitle={2023 IEEE international conference on robotics and automation (ICRA)},
  pages={2774--2781},
  year={2023},
  organization={IEEE}
}

@article{chai2022deep,
  title={Deep learning-based trajectory planning and control for autonomous ground vehicle parking maneuver},
  author={Chai, Runqi and Liu, Derong and Liu, Tianhao and Tsourdos, Antonios and Xia, Yuanqing and Chai, Senchun},
  journal={IEEE Transactions on Automation Science and Engineering},
  volume={20},
  number={3},
  pages={1633--1647},
  year={2022},
  publisher={IEEE}
}

@article{huang2022bevdet4d,
  title={Bevdet4d: Exploit temporal cues in multi-camera 3d object detection},
  author={Huang, Junjie and Huang, Guan},
  journal={arXiv preprint arXiv:2203.17054},
  year={2022}
}

@article{chai2018two,
  title={Two-stage trajectory optimization for autonomous ground vehicles parking maneuver},
  author={Chai, Runqi and Tsourdos, Antonios and Savvaris, Al and Chai, Senchun and Xia, Yuanqing},
  journal={IEEE Transactions on Industrial Informatics},
  volume={15},
  number={7},
  pages={3899--3909},
  year={2018},
  publisher={IEEE}
}

@inproceedings{li2023bevdepth,
  title={Bevdepth: Acquisition of reliable depth for multi-view 3d object detection},
  author={Li, Yinhao and Ge, Zheng and Yu, Guanyi and Yang, Jinrong and Wang, Zengran and Shi, Yukang and Sun, Jianjian and Li, Zeming},
  booktitle={Proceedings of the AAAI conference on artificial intelligence},
  volume={37},
  pages={1477--1485},
  year={2023}
}

@article{li2024fast,
  title={Fast-BEV: A fast and strong bird's-eye view perception baseline},
  author={Li, Yangguang and Huang, Bin and Chen, Zeren and Cui, Yufeng and Liang, Feng and Shen, Mingzhu and Liu, Fenggang and Xie, Enze and Sheng, Lu and Ouyang, Wanli and others},
  journal={IEEE Transactions on Pattern Analysis and Machine Intelligence},
  year={2024},
  publisher={IEEE}
}

@article{chitta2022transfuser,
  title={Transfuser: Imitation with transformer-based sensor fusion for autonomous driving},
  author={Chitta, Kashyap and Prakash, Aditya and Jaeger, Bernhard and Yu, Zehao and Renz, Katrin and Geiger, Andreas},
  journal={IEEE transactions on pattern analysis and machine intelligence},
  volume={45},
  number={11},
  pages={12878--12895},
  year={2022},
  publisher={IEEE}
}

@inproceedings{jaeger2023hidden,
  title={Hidden biases of end-to-end driving models},
  author={Jaeger, Bernhard and Chitta, Kashyap and Geiger, Andreas},
  booktitle={Proceedings of the IEEE/CVF International Conference on Computer Vision},
  pages={8240--8249},
  year={2023}
}

@inproceedings{codevilla2018end,
  title={End-to-end driving via conditional imitation learning},
  author={Codevilla, Felipe and M{\"u}ller, Matthias and L{\'o}pez, Antonio and Koltun, Vladlen and Dosovitskiy, Alexey},
  booktitle={2018 IEEE international conference on robotics and automation (ICRA)},
  pages={4693--4700},
  year={2018},
  organization={IEEE}
}

@article{shamsoshoara2024swaptransformer,
  title={Swaptransformer: highway overtaking tactical planner model via imitation learning on osha dataset},
  author={Shamsoshoara, Alireza and Salih, Safin B and Aghazadeh, Pedram},
  journal={IEEE Access},
  year={2024},
  publisher={IEEE}
}

@inproceedings{mu2024pix2planning,
  title={Pix2Planning: End-to-End Planning by Vision-language Model for Autonomous Driving on Carla Simulator},
  author={Mu, Xiangru and Qin, Tong and Zhang, Songan and Xu, Chunjing and Yang, Ming},
  booktitle={2024 IEEE Intelligent Vehicles Symposium (IV)},
  pages={2383--2390},
  year={2024},
  organization={IEEE}
}

@inproceedings{lyu2024sensor,
  title={Sensor Fusion and Motion Planning with Unified Bird’s-Eye View Representation for End-to-end Autonomous Driving},
  author={Lyu, Yuandong and Tan, Xiaojun and Yu, Ze and Fan, Zhengping},
  booktitle={2024 International Joint Conference on Neural Networks (IJCNN)},
  pages={1--8},
  year={2024},
  organization={IEEE}
}

@article{wu2022trajectory,
  title={Trajectory-guided control prediction for end-to-end autonomous driving: A simple yet strong baseline},
  author={Wu, Penghao and Jia, Xiaosong and Chen, Li and Yan, Junchi and Li, Hongyang and Qiao, Yu},
  journal={Advances in Neural Information Processing Systems},
  volume={35},
  pages={6119--6132},
  year={2022}
}

@inproceedings{shen2020parkpredict,
  title={Parkpredict: Motion and intent prediction of vehicles in parking lots},
  author={Shen, Xu and Batkovic, Ivo and Govindarajan, Vijay and Falcone, Paolo and Darrell, Trevor and Borrelli, Francesco},
  booktitle={2020 IEEE Intelligent Vehicles Symposium (IV)},
  pages={1170--1175},
  year={2020},
  organization={IEEE}
}

@inproceedings{shen2022parkpredict+,
  title={Parkpredict+: Multimodal intent and motion prediction for vehicles in parking lots with cnn and transformer},
  author={Shen, Xu and Lacayo, Matthew and Guggilla, Nidhir and Borrelli, Francesco},
  booktitle={2022 IEEE 25th International Conference on Intelligent Transportation Systems (ITSC)},
  pages={3999--4004},
  year={2022},
  organization={IEEE}
}

@inproceedings{yang2024e2e,
  title={E2e parking: Autonomous parking by the end-to-end neural network on the carla simulator},
  author={Yang, Yunfan and Chen, Denglong and Qin, Tong and Mu, Xiangru and Xu, Chunjing and Yang, Ming},
  booktitle={2024 IEEE Intelligent Vehicles Symposium (IV)},
  pages={2375--2382},
  year={2024},
  organization={IEEE}
}

@inproceedings{li2024parkinge2e,
  title={ParkingE2E: Camera-based End-to-end Parking Network, from Images to Planning},
  author={Li, Changze and Ji, Ziheng and Chen, Zhe and Qin, Tong and Yang, Ming},
  booktitle={2024 IEEE/RSJ International Conference on Intelligent Robots and Systems (IROS)},
  pages={13206--13212},
  year={2024},
  organization={IEEE}
}

@inproceedings{tan2019efficientnet,
  title={Efficientnet: Rethinking model scaling for convolutional neural networks},
  author={Tan, Mingxing and Le, Quoc},
  booktitle={International conference on machine learning},
  pages={6105--6114},
  year={2019},
  organization={PMLR}
}

\end{document}